\documentclass[letterpaper, 10 pt, conference]{ieeeconf}  
\usepackage{tcolorbox}
\usepackage{xcolor}
\usepackage{booktabs}
\usepackage{multicol}
\usepackage{multirow}
\usepackage{amsmath}
\usepackage{url}
\usepackage{booktabs}
\usepackage{tabularx} 
\usepackage{makecell} 
\usepackage{adjustbox}
\usepackage{makecell}
\usepackage{makecell}
 
\newcommand{\Judy}[1]{\textcolor{black}{#1}}
\newcommand{\ZB}[1]{\textcolor{black}{#1}}
\newcommand{\TBD}[1]{\textcolor{black}{#1}} 

\usepackage{soul}

\def\ie{{\textit{i.e.}}}
\def\eg{{\textit{e.g.}}}

\IEEEoverridecommandlockouts                              

\title{\LARGE \bf
Can LLMs Prove Robotic Path Planning Optimality? A Benchmark for Research-Level Algorithm Verification 
}

\author{
 \textbf{Zhengbang Yang\textsuperscript{1}},
 \textbf{Md. Tasin Tazwar\textsuperscript{2}},
 \textbf{Minghan Wei\textsuperscript{2}},
 \textbf{Zhuangdi Zhu\textsuperscript{1}}
 \thanks{\textsuperscript{1}George Mason University. \textsuperscript{2}Florida Atlantic University.}
 \thanks{Email: \{zyang30,zzhu24\}@gmu.edu, \{mtazwar2023,weim\}@fau.edu. }
 \thanks{This work was supported by NSF Award No. 2452203. }
 \thanks{Source code: https://github.com/Zhengbang-Yang/RoboticsBench}
}

\begin{document}

\maketitle
\thispagestyle{empty}
\pagestyle{empty}

\begin{abstract}
Robotic path planning problems are often NP-hard, and practical solutions typically rely on approximation algorithms with provable performance guarantees for general cases. While designing such algorithms is challenging, formally proving their approximation optimality is even more demanding, which requires domain-specific geometric insights and multi-step mathematical reasoning over complex operational constraints.
Recent Large Language Models (LLMs) have demonstrated strong performance on mathematical reasoning benchmarks, yet their ability to assist with research-level optimality proofs in robotic path planning remains under-explored.
In this work, we introduce the first benchmark for evaluating LLMs on approximation-ratio proofs of robotic path planning algorithms. The benchmark consists of 34 research-grade proof tasks spanning diverse planning problem types and complexity levels, each requiring structured reasoning over algorithm descriptions, problem constraints, and theoretical guarantees.
Our evaluation of state-of-the-art proprietary and open-source LLMs reveals that even the strongest models struggle to produce fully valid proofs without external domain knowledge. 
However, providing LLMs with task-specific in-context lemmas substantially improves reasoning quality, a factor that is more effective than generic chain-of-thought prompting or supplying the ground-truth approximation ratio as posterior knowledge. We further provide fine-grained error analysis to characterize common logical failures and hallucinations, and demonstrate how each error type can be mitigated through targeted context augmentation. 
\end{abstract}


\section{Introduction}
Robotic path planning problems are usually NP-hard~\cite{DUAN2024101576, BAZGANkvrp}, and this field has long relied on approximation algorithms as practical solutions. Even marginal improvements in approximation ratios can yield substantial real-world benefits in deployment time, energy consumption, and mission reliability in varying high-stake domains~\cite{DUAN2024101576, BAZGANkvrp}. Therefore, the quality of proposed algorithms plays significant role. Yet while proposing a new planning algorithm is relatively tractable, formally proving its approximation optimality is far more demanding, often requiring human experts days or weeks of careful mathematical and geometric reasoning.

The recent emergence of large language models (LLMs) with strong performance in mathematical reasoning~\cite{shao2024deepseekmathpushinglimitsmathematical,gao2025omnimath,he-etal-2024-olympiadbench,balunovic2025matharena} raises a natural question: \textit{can LLMs expedite  the formal verification of approximation optimality for robotic path planning algorithms?} Such a capability would lay the groundwork for automated algorithm validation, and ultimately for the cost-effective discovery of new planning algorithms with stronger optimality guarantees.
One might expect LLMs to transfer their mathematical competency to this domain, particularly given their advanced performance on math reasoning benchmarks such as \TBD{GMS8K}~\cite{cobbe2021trainingverifierssolvemath} and MATH~\cite{hendrycks2021measuring}. 
However, the task we investigate is fundamentally distinct, where the planning optimality proofs operate at a research level, involving complex optimization constraints arising from operational requirements, such as cost budgets and terrain or operational restrictions~\cite{Wei_Isler_2018,KATOH20062335}. 
They demand task-specific knowledge of domain lemmas and theorems, and  reasoning over long problem formulations that jointly describe the algorithm, the environment, and the optimality objective. 
\TBD{Furthermore, a valid proof typically requires multi-step deduction to connect concrete task descriptions to abstract mathematical conditions. For instance, proving the 3/2-approximation guarantee for the metric Traveling Salesman Problem (TSP) requires mapping the tour-planning objective to combinatorial structures (minimum spanning trees (MST), odd-degree vertex sets, and minimum-weight perfect matchings) and then chaining multiple lemmas, \eg, $\mathrm{MST}\le \mathrm{OPT}$, $\mathrm{MWPM}\le \mathrm{OPT}/2$, and triangle-inequality shortcutting to convert an Eulerian walk into a Hamiltonian tour without increasing cost.} 
These characteristics place our investigation well beyond the scope of standard math reasoning.

To systematically study this challenge, we introduce a benchmark comprising \Judy{34} robotic constrained planning problems from \Judy{11} peer-reviewed, published robotic papers. Each instance includes the original problem formulation, algorithm description, and formal performance guarantee as presented in the source work. The tasks are curated from representative constrained path-planning formulations motivated by real robotic deployment settings, including energy-aware navigation, perception-constrained routing, time-critical missions, and multi-robot workload balancing. Each selected paper establishes formal approximation guarantees, enabling us to assess whether LLMs can reason about algorithmic performance bounds rather than merely solving isolated planning instances.
\Judy{
\textbf{\textit{Our evaluation reveals several key findings:}}
(1) Current LLMs differ substantially in performance, yet even the strongest models struggle to produce valid optimality proofs without given expert domain knowledge (setting 1, Sec \ref{subsec:main-result}).
(2) Augmenting prompts with task-relevant in-context knowledge, specifically, lemmas and prior knowledge tailored to the target problem, leads to notable improvements in proof accuracy.
(3) Providing the ground-truth approximation ratio as posterior knowledge also improves reasoning quality, though it contributes less than in-context lemmas.
(4) Generic chain-of-thought (CoT)~\cite{CoT} or deduction-inducing prompting alone does not meaningfully reduce task difficulty, though it provides complementary benefit when combined with posterior knowledge.
(5) Notably, we find that open-source models such as Qwen 3.5~\cite{qwen3.5} respond strongly to in-context augmentation, with performance gains that can surpass those achieved by counterpart LLMs.}
We further provide a fine-grained error analysis (Sec~\ref{subsec:error-analysis}) to characterize where LLMs typically fail and how each error type can be mitigated. 
Our benchmark also introduces a granular evaluation metric that jointly assesses the correctness of the final proof, the quality of the reasoning chain, and the alignment with ground-truth proofs from the original research papers. We releases an automated pipeline for extracting and formalizing path planning algorithms from research papers to facilitate future work in this domain.

\section{Related Work}

    \noindent \textbf{\textit{LLM assisted path planning:}}
Work along this line explores using LLMs directly as planning agents or algorithm executors.
NLGraph benchmark~\cite{wang2023can} demonstrated that LLMs possess preliminary abilities to trace shortest paths on simple classic topologies.
LLM+P ~\cite{liu2023llmpempoweringlargelanguage} integrated language models with classical planners, utilizing the LLM to translate natural language problem descriptions into formal specifications that symbolic solvers can process.
LLM-A* ~\cite{meng-etal-2024-llm} integrated language models into the A* search procedure to heuristically navigate planning spaces.
Similarly, GraphArena ~\cite{tang2025grapharena} benchmarked LLMs on their ability to act as algorithmic executors for graph computation tasks, testing their step-by-step traversal on complex routing challenges like the Traveling Salesman Problem. 
In practical domains, frameworks like TravelPlanner ~\cite{TravelPlanner} and Personal Travel Solver ~\cite{PersonalTravelSolver} deployed LLMs as autonomous solvers for real-world routing and planning instances, assessing their ability to satisfy multifaceted constraints.
They evaluate LLMs as \textit{solvers} of planning instances rather than investigating whether LLMs can reason about the optimality of proposed algorithms.

\noindent \textbf{\textit{LLM on scientific or mathematical reasoning:}} A parallel thread evaluates LLMs on scientific and mathematical reasoning more broadly.
%
%
Datasets such as GMS8K~\cite{cobbe2021trainingverifierssolvemath}, MATH~\cite{hendrycks2021measuring}, PRM800K~\cite{lightman2023let}  provided mathematical problems ranging from elementary to undergraduate-level difficulty. State-of-the-art LLMs tend to saturate on these math benchmarks with (near-) expert performance. More advanced benchmarks include ARB~\cite{sawada2024arb}, OmniMATH~\cite{gao2025omnimath}, Olympiadbench~\cite{he-etal-2024-olympiadbench}, Matharena~\cite{balunovic2025matharena}.
%
%
They target general mathematical competitions where problems are mostly self-contained and drawn from established Olympiad corpora.
While FrontierMath ~\cite{glazer2025frontiermathbenchmarkevaluatingadvanced} remains a formidable challenge for state-of-the-art models, it merely evaluates final answers rather than demanding complete proofs or granular reasoning steps. Furthermore, its closed-source nature limits community-driven research, both of which are essential for advancing the field.
Some other work focused on training or inference time \textit{optimization} instead of \textit{benchmarking} to improve the geometry or mathematical question-solving ability of LLMs \cite{chervonyi2025goldmedalistperformancesolvingolympiad,shao2024deepseekmathpushinglimitsmathematical}. 

\noindent \textit{\textbf{LLM for inequality proof:}} Our work draws a close connection to those that investigate LLM's ability to prove inequalities. 
IneqMath~\cite{sheng2025solving} evaluated LLMs on Olympiad-level inequality proving with a neutral language formulation and step-wise judges. 
Open Proof Corpus ~\cite{dekoninck2026the} provided a dataset of human-evaluated LLM-generated Olympiad-level proofs. However, the complexity of these existing works falls short of research-level standards, and they lack a detailed analysis of the errors in model-generated proofs.
Our work differs from the above efforts along three critical dimensions: \textit{domain specificity}, \textit{task type and complexity}, and \textit{evaluation granularity}.
It targets research-level proofs of approximation optimality for robotic path-planning algorithms, which require simultaneous mastery of domain-specific lemmas, long-context algorithm descriptions with operational constraints, and multi-step deductive reasoning, all of which are lacking in existing LLM reasoning benchmarks.

\noindent \textit{\textbf{LLM as Judge for scientific reasoning:}} As LLMs have become more powerful in reasoning, a common evaluation strategy is to reuse them as automatic graders, reducing the cost of assessing open-ended generations while retaining meaningful signal about quality ~\cite{Zheng2023Judging,gu2024survey,chevalier2024language}. Building on this line, multiple studies have explored LLM-based judging for mathematical solutions and proofs, reporting that with appropriately scoped rubrics, LLM graders can be sufficiently consistent for certain proof-style tasks and can also support feedback-oriented educational use cases ~\cite{sheng2025solving,zhao2025autograding,zhao2025language}.
Specifically, for inequality proofs, IneqMath~\cite{sheng2025solving} validated that LLM judging can echo human judgments when the evaluation is decomposed appropriately. In parallel, OPC~\cite{dekoninck2026the} shows that top LLM judges achieve near-human judging accuracy on proof correctness via a large-scale study, reinforcing that LLM-as-judge can be highly consistent with expert human assessment in proof settings. 
\Judy{Built upon this prior work, we adopt LLM-as-Judge for the first-round evaluation and error analysis, followed by human researcher verifications.}

\section{Dataset Composition}

\subsection{Task Definition}  
\begin{figure}
    \centering
    \includegraphics[width=0.8\linewidth]{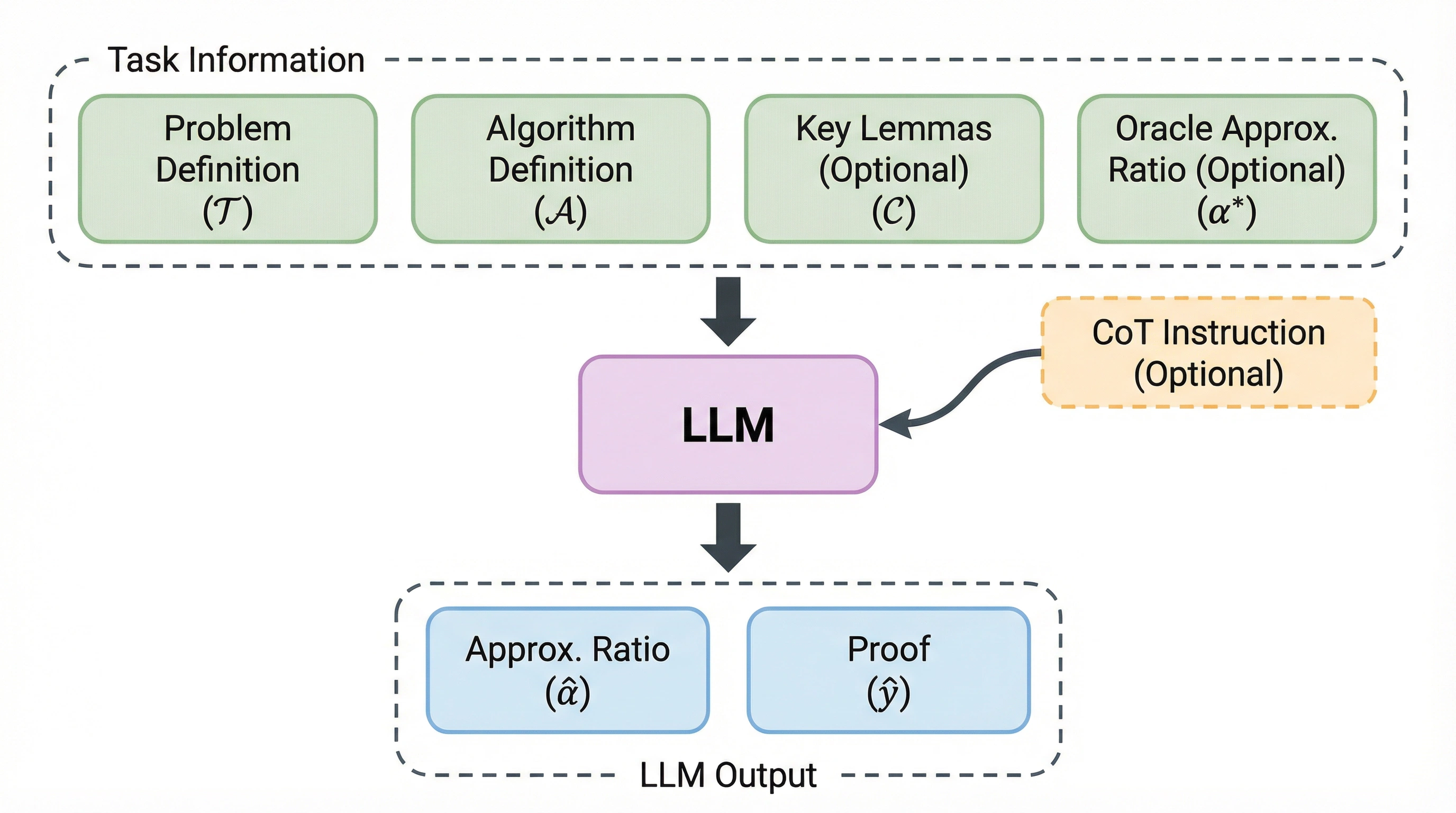}
    \vspace{-0.1in}
    \caption{LLM proof generation pipeline. The LLM receives a path-planning problem $\mathcal{T} $ and algorithm $\mathcal{A} $, along with \textit{\textbf{optional}} in-context lemmas and theorems $\mathcal{C} $. It is then prompted (\textit{\textbf{optionally)}} with a \textit{Chain-of-Thought} style instruction) to generate the approximation ratio $\hat{\alpha}$ and a corresponding proof $\hat{y} $, both evaluated against ground-truth references. When ground-truth $\alpha^*$ is also provided as part of the input, the task reduces to generating reasoning $\hat{y} $ proving how the given $\alpha^*$ is derived. }
    \label{fig:pipeline}
\end{figure}

Each task in our benchmark requires establishing the theoretical performance guarantees of a provided algorithm $\mathcal{A}$ for a robotics path planning problem $\mathcal{T}$, namely, proving that the cost of the provided algorithm is at most $\alpha$ times the cost of the optimal solution, where $\alpha$ is referred to as the approximation ratio. 
As illustrated in Figure~\ref{fig:pipeline}, each task consists of a problem description $\mathcal{T}$, the given algorithm $\mathcal{A}$ to sovle $\mathcal{T}$, an \textit{\textbf{optional}} set of in-context lemmas $\mathcal{C}$, and an \textit{\textbf{optional}} posterior approximation ratio $\alpha$, which together form the input provided to the LLM under evaluation; the model is then asked to produce a proof $\hat{y}$. 
We also provide an \textit{\textbf{optional}} \textit{Chain-of-Thought} style reasoning template to assist the LLM in generating legit proofs. 
Figure \ref{fig:pipeline} overviews the proof generation process of the LLMs under investigation.

\TBD{For comparison, Figure \ref{fig:comparsion} compares a mathematical inequality proof drawn from an existing benchmark~\cite{sheng2025solving} and proof of robotic path planning optimality.}
Such inequality proofs are typically simpler in structure, relying on standard lemmas and algebraic manipulation with few operational constraints. In contrast, path planning optimality proofs are multi-lemma, long-context, and constraint-coupled.

\begin{figure*}[t]
\vspace{+0.1in}
\centering
\begin{tcolorbox}[
  colback=gray!12,
  colframe=black!70,
  boxrule=0.8pt,
  arc=2.5mm,
  left=6pt,
  right=6pt,
  top=4pt,
  bottom=6pt,
  title=\textbf{Task Example: Path Planning Optimality Proof \textit{vs.} Math Inequality Proof},
  coltitle=white,
  colbacktitle=black,
  fonttitle=\bfseries,
  fontupper=\normalfont
]
\begin{small}
\textbf{Example-1: Math Inequality Proof Task}
\begin{itemize}
    \item \textit{Problem formulation}: Find the maximal constant $C$ such that for all real $a,b,c$,
  \[\sqrt{a^2+(1-b)^2}+\sqrt{b^2+(1-c)^2}+\sqrt{c^2+(1-a)^2}\ge C.\]
  \textit{Helpful Lemmas}: Minkowsky’s Inequality Theorem: ...
  \item \textit{Key Proof Steps}:
    \begin{enumerate}
      \item Apply Minkowski (or a norm/triangle inequality) to lower bound a sum of square-roots by a single square-root of aggregated terms.
      \item Reduce the expression to a 1D quadratic in $x=a+b+c$ and take the minimum.
      \item Conclude the best constant $C$ and give a tightness witness (equality case).
    \end{enumerate}
\end{itemize}

\medskip

\textbf{Example-2: Path Planning Optimality Proof Task} \textit{Christofides Algorithm for TSP (as a simplest case)}:
\begin{itemize}
    \item \textit{Problem formulation:}  
        For any metric TSP instance with optimal tour cost $\mathrm{OPT}$, prove the approximation ratio of the Christofides algorithm.
        \begin{itemize}
            \item TSP problem definition: ...
            \item \textbf{Constraints} of the problem: Hamiltonian cycle $\cdots$
            \item Christofides algorithm definition: ...
            \item Helpful Lemmas (\textit{\textbf{Optional}} Information): Lemma 1 (MST bound): ..., Lemma 2 (matching bound): ..., ...
            \item Ground truth approximation ratio $\alpha$ (\textit{\textbf{Optional}} Information)
        \end{itemize}
    
    \item \textit{Key Proof Steps}: 
        \begin{enumerate}
          \item \textbf{Algorithm structure:} build an MST $T$; let $O$ be odd-degree vertices in $T$; compute a minimum-weight perfect matching $M$ on $O$; combine $T\cup M$ to get an Eulerian multigraph; take an Euler tour and \emph{shortcut} repeated vertices to obtain a Hamiltonian cycle.
          \item \textbf{Lemma 1 (MST bound):} $\mathrm{cost}(T)\le \mathrm{OPT}$ (MST no more expensive than removing one edge from an optimal tour).
          \item \textbf{Lemma 2 (matching bound):} $\mathrm{cost}(M)\le \tfrac{1}{2}\mathrm{OPT}$ (odd-degree set has even size; the optimal tour restricted to $O$ can be decomposed into two perfect matchings, so the minimum one is at most half).
          \item \textbf{Lemma 3 (Euler $\rightarrow$ tour feasibility):} $T\cup M$ is Eulerian; the Euler walk can be shortcut to a Hamiltonian cycle without increasing cost \emph{only because} the instance is metric (triangle inequality).
          \item \textbf{Combine:} $\mathrm{cost}(\mathcal{A})\le \mathrm{cost}(T)+\mathrm{cost}(M)\le \mathrm{OPT}+\tfrac{1}{2}\mathrm{OPT}=\tfrac{3}{2}\mathrm{OPT}$.
        \end{enumerate}
\end{itemize}

 \end{small}
\end{tcolorbox}
\vspace{-0.1in}
\caption{An Example Comparison of a Path Planning Optimality Proof \textit{vs.} a Generic Math Inequality Proof.}
\label{fig:comparsion}
\end{figure*}

\subsection{Task Distribution}
\ZB{We collected $34$ robotic path planning research methods from \TBD{11} papers which cover research-level robotic path-planning and routing problems that require jointly reasoning about (i) operational constraints (\eg, energy/distance budgets, charging, pickups/deliveries, regret or lateness costs), (ii) structured environments (\eg, grid/polygon/contour-connected spaces, tree metrics, general metrics, and 3D visibility-cone geometry), and (iii) approximation optimality proofs that combine algorithmic decomposition with non-trivial inequalities and geometric arguments. 
Concretely, the collected methods cluster into: (A) energy-constrained coverage path planning (unit grid/polygon/contour-connected environments with charging constraints), 
(B) distance/capacity-constrained vehicle routing and its variants (DVRP~\cite{DVRP}, CVRP-on-tree~\cite{CVRP}, kTSP~\cite{BAZGANkvrp,4567906} objectives such as min–max and min–sum, and regret-bounded routing), and (C) geometric neighborhood touring such as Cone TSPN~\cite{Cone-TSPN} in 3D (visibility cones, intersecting vs disjoint neighborhoods). 
These categories emphasize domain specificity (robotic coverage + routing constraints), research-level task complexity (long-context algorithm descriptions and multi-lemma proofs), and evaluation granularity (proof-step correctness rather than final answers alone). } 
%


\subsection{Data Preparation}
\noindent \textbf{Paper Data Collection.}
\ZB{We collected the raw corpus of papers through an expert-driven selection process: domain experts in robotic path planning actively searched for and shortlisted suitable research papers that are representative of real-world planning and routing settings, and that contain non-trivial algorithmic designs and approximation-optimality proofs appropriate for inclusion in our benchmark.}

\noindent \textbf{Data Extraction and Standardization.}
\ZB{To scale and standardize the paper, especially math notations and tables, we designed an automated extraction pipeline that uses GPT-5.2 as an agent to process of the PDF files of papers. Given each paper PDF, the pipeline first converts the contents into a structured LaTeX document; we also developed a conversion pipeline that explicitly emphasizes faithful reproduction of mathematical expressions and tables to maximize the data quality. 
Based on the resulting LaTeX document, the agent then extracts (i) the formal problem definition(s) $\mathcal{T}$, 
(ii) the algorithm definition $\mathcal{A}$,
(iii) the ground-truth approximation ratio $\alpha$ and its corresponding proof, and (iv) the key lemmas invoked by the ground truth proof $\mathcal{C}$. 
This structured representation supports downstream evaluation in various settings. 
}

\noindent \textbf{Problem and Algorithm Anonymization.} Notably, we observed that providing the name of a classic algorithm or problem, such as the \textit{k}-TSP~\cite{4567906}, can illicitly prompt an LLM to produce a fluent proof, \TBD{likely due to \textit{data contamination}}, where the algorithm name serves as a retrieval cue that triggers verbatim reproduction of memorized content rather than genuine reasoning. To prevent this, we have anonymized all problem and algorithm names in our benchmark.

\noindent \textbf{Human verification.} 4 human annotators (faculty and graduate students) have manually verified the quality of extracted data from each research paper, and the evaluation quality of the LLM-as-judge, which helped refine the evaluation instructions prompt iteratively.

\section{Evaluation}

\subsection{Experimental Settings}
\noindent \textbf{Task Difficulty:} We evaluated LLMs under \Judy{four} settings of varying information availability.
In \textbf{\textit{Setting-1}} (no context), the LLM receives only the task definition and algorithm description $\mathcal{A}$.
In \textbf{\textit{Setting-2}} (context-augmented), the LLM additionally receives context $\mathcal{C}$ containing key lemmas extracted from the ground-truth research paper. 
In \textbf{\textit{Setting-3}} (posterior-augmented), the LLM is provided with the ground-truth approximation ratio $\alpha$ but no supporting lemmas. Specifically, setting-3 allows us to investigate whether this posterior information alone can guide the model to reconstruct a correct reasoning path in retrospect.
\ZB{In \textbf{\textit{Setting-4}} (context and posterior-augmented), the LLM is provided with both context $\mathcal{C}$ and the ground-truth approximation ratio $\alpha$.}
%


\begin{table*}[t]
\centering
\small
\vspace{+0.1in}
\setlength{\tabcolsep}{1pt}
\renewcommand{\arraystretch}{1.1}
\caption{\textbf{Main results under four evaluation settings:} scores (1--10) for \textit{Final Answer}, \textit{Reasoning}, and \textit{Relevance}, evaluated \textit{with} and \textit{without} a crafted generic reasoning template. Final-answer scores are \Judy{omitted} (indicated with '-') for Setting-3 and 4, since the oracle final approximation ratio is given in these settings. In case of a consistent timeout issue in LLM reasoning, we assigned 1 (the lowest score) for each evaluation metric. \ZB{Success Rate is the percentage ratio of success, where success is defined as a score of $\geq 7 $ for both the final answer and the reasoning steps in Setting-1 and 2, and a reasoning score $\geq 7 $ in Setting-3 and 4.}}
\resizebox{0.9\textwidth}{!}{
\begin{tabularx}{\textwidth}{l *{7}{>{\centering\arraybackslash}X} >{\centering\arraybackslash}X}
\toprule
\textbf{Model} &
\textbf{Final Answer $\uparrow$} & \textbf{Final Answer $\uparrow$} &
\textbf{Reasoning $\uparrow$} & \textbf{Reasoning $\uparrow$} &
\textbf{Relevance $\uparrow$} & \textbf{Relevance $\uparrow$} &
\textbf{Success Rate $\uparrow$}\\
 &
\textbf{(w/o CoT)} & \textbf{(w/ CoT)} &
\textbf{(w/o CoT)} & \textbf{(w/ CoT)} &
\textbf{(w/o CoT)} & \textbf{(w/ CoT)} & \textbf{ (\%) (w/o CoT)} \\
\midrule

\multicolumn{8}{c}{\textbf{Setting-1} (\textit{no-context)}} \\
\midrule
GPT 5.2       & \textbf{4.15} & 3.35 & \textbf{4.62} & \textbf{4.47} & \textbf{4.88} & 4.12 & \textbf{26.47}\\
Gemini 3 Pro  & 3.71 & \textbf{3.59} & 3.47 & 3.53 & 3.94 & \textbf{4.18} & 11.76 \\
Grok 4.1      & 2.88 & 2.76 & 3.24 & 3.00 & 3.59 & 3.41 & 14.71 \\
Qwen3.5       & 3.53 & 3.03 & 3.32 & 3.53 & 4.09 & 3.85 & 8.82 \\
\midrule

\multicolumn{8}{c}{\textbf{Setting-2} \textit{(with context)}}  \\
\midrule
GPT 5.2       & 4.44 & 4.32 & 4.94 & 4.82 & 4.91 & 4.88 & 32.35 \\
Gemini 3 Pro  & 5.62 & 5.50 & 5.06 & 5.03 & 6.32 & 6.44 & 35.29 \\
Grok 4.1      & 4.62 & 4.56 & 4.26 & 4.09 & 5.56 & 5.41 & 29.41 \\
Qwen3.5       & \textbf{5.97} & \textbf{6.06} & \textbf{5.32} & \textbf{5.35} & \textbf{6.68} & \textbf{6.35} & \textbf{44.12} \\
\midrule

\multicolumn{8}{c}{\textbf{Setting-3} \textit{(with oracle ratio)}} \\
\midrule
GPT 5.2       & -- & -- & 4.04 & 4.22 & 4.46 & 4.26 & \textbf{32.35} \\
Gemini 3 Pro  & -- & -- & \textbf{4.68} & \textbf{4.87} & 5.82 & \textbf{6.34} & \textbf{32.35} \\
Grok 4.1      & -- & -- & 3.59 & 4.10 & 4.63 & 5.35 & 17.65 \\
Qwen3.5       & -- & -- & 4.38 & 4.50 & \textbf{5.97} & 5.65 & 23.53 \\
\midrule

\multicolumn{8}{c}{\textbf{Setting-4} \textit{(with context and oracle ratio)}} \\
\midrule
GPT 5.2       & -- & -- & 5.26 & 5.01 & 5.59 & 5.50 & \textbf{50.00} \\
Gemini 3 Pro  & -- & -- & \textbf{5.46} & \textbf{6.32} & \textbf{7.21} & \textbf{7.56} & 38.24 \\
Grok 4.1      & -- & -- & 4.66 & 5.24 & 5.85 & 6.44 & 41.18 \\
Qwen3.5       & -- & -- & 5.15 & 5.25 & 7.12 & 7.01 & 32.35 \\
\bottomrule
\end{tabularx}
}
\label{tab:scores}
\end{table*}

\noindent \textbf{Models Under Evaluation}  include Grok 4.1~\cite{grok4.1}, Gemini 3 Pro~\cite{Gemini3Pro}, GPT 5.2~\cite{GPT5.2}, and Qwen3.5~\cite{qwen3.5} (397B-A17B), covering a broad slice of today’s state-of-the-art LLMs across both closed-source commercial systems and strong open-source releases. For all models, we enabled their reasoning mode to reflect best-effort performance on these proof-intensive questions. Claude 4.6 was also tested, but it frequently timed out on these questions, making its results non-informative and therefore excluded from further analysis. 

\subsection{Evaluation Metrics} 
We used the ground-truth proof of each task to guide an LLM-as-judge (\TBD{GPT-5.2}) in evaluating the model-generated proof $\hat{y}$ against the ground-truth $y$ along the following three dimensions, each scored on a scale of 1 to 10: 

\noindent \textit{\textbf{Final answer}} measures how accurately the predicted approximation ratio matches the ground-truth, accounting for exactness of constants, asymptotic order, and parameter dependence. 
A score of 9-10 indicates the ratio is exact or differs only in minor presentation; 6-8 reflects the correct asymptotic order with nontrivial mismatches in constants or parameter dependence; and scores below 6 indicate an incorrect approximation ratio.

\noindent \textit{\textbf{The reasoning correctness score}} assesses the logical validity of the derivation, which penalizes unjustified steps, incorrect inequalities, and missing feasibility arguments. 
A score of 9-10 indicates a fully valid proof with all key steps justified; 6-8 reflects a proof whose main strategy is sound but contains repairable gaps; and scores below 6 indicate significant or fatal logical errors.

\noindent \textit{\textbf{The proof relevance score}} captures how well the generated proof is grounded in the provided problem context. It rewards precise use of supplied lemmas and definitions while penalizing reliance on invented or irrelevant reasoning. 
A score of 9-10 indicates the proof directly and correctly references the provided context; 6-8 reflects partial alignment with some loosely connected segments; and scores below 6 indicate weak alignment or reliance on fabricated steps.

\subsection{Main Results}\label{subsec:main-result}
State-of-the-art LLMs broadly struggle with algorithmic proof reasoning: without external domain knowledge  (Table \ref{tab:scores}, Setting-1), only up to \TBD{26.47\%} of the path planning algorithms in our benchmark are successfully addressed, where \textbf{\textit{success}} is defined as a score of $\geq 7 $ for both the final answer and the reasoning steps in Setting-1 and 2, and a reasoning score $\geq 7 $ in Setting-3 and 4.
Beyond this headline finding, our benchmark also surfaces answers to several intriguing questions.


\noindent \textbf{\textit{Can providing external information help LLMs get back on the right track?}} Yes, we found that both providing in-context lemma (Setting-2) and an oracle posterior (Setting-3) help LLMs in terms of general performance, where in-context lemma (Setting-2) plays a more important role than providing posterior or CoT. This is supported by results in Table ~\ref{tab:scores}, \Judy{where the best performance in Setting-2 consistently surpasses  those in Setting-3.}

\noindent \ZB{\noindent \textbf{\textit{Is oracle posterior information useful when in-context lemmas are already provided?}} Yes. While posterior-only guidance (Setting-3) yields limited benefit in recovering a correct proof strategy, Setting-4 shows that $\alpha$ becomes helpful \emph{once the model is anchored by domain lemmas $\mathcal{C}$}. Concretely, when supplied with both $\mathcal{C}$ and $\alpha$, models achieve higher reasoning-correctness and relevance scores than with $\mathcal{C}$ alone (Setting-2), indicating that the oracle ratio can serve as an effective \emph{consistency check} that nudges the model toward the intended proof target and discourages drifting to alternative ratios. This suggests that posterior information is not a substitute for domain knowledge; instead, it acts as a \emph{complementary signal} that refines and stabilizes the derivation after the proof skeleton has been grounded by the provided lemmas.}

\noindent \Judy{\textit{\textbf{Does generic CoT prompting help LLM achieve better performance?}}
Comparing results with and without CoT in Table~\ref{tab:scores}, generic deduction-inducing prompting alone does not meaningfully reduce task difficulty across settings. However, when combined with posterior knowledge (Settings-3 and 4), CoT provides a complementary benefit on the \textit{Reasoning} score across most LLMs. This indicates that while CoT is insufficient as a standalone strategy, it can reinforce structured reasoning when the model is already anchored by correct posterior information.
}

\noindent \textbf{\textit{Which LLMs are most capable on these tasks?}} In the strictest setting, \ie\ no in-context lemmas, no posterior, proprietary models lead, with GPT 5.2 outperforming open-source alternatives ( GPT 5.2 $>$ Gemini 3 Pro $>$ Qwen 3.5 $>$ Grok 4.1 ), as shown in Table ~\ref{tab:scores}. Notably, however, the open-source model Qwen 3.5 shows great potential when task-tailored in-context information is provided (Setting-2), which outperforms other LLM counterparts, suggesting the strong latent capacity of this open-sourced model that benefits from domain-specific guidance. 

\begin{figure*}[htbp!]
    \centering
    \includegraphics[width=0.75\linewidth]{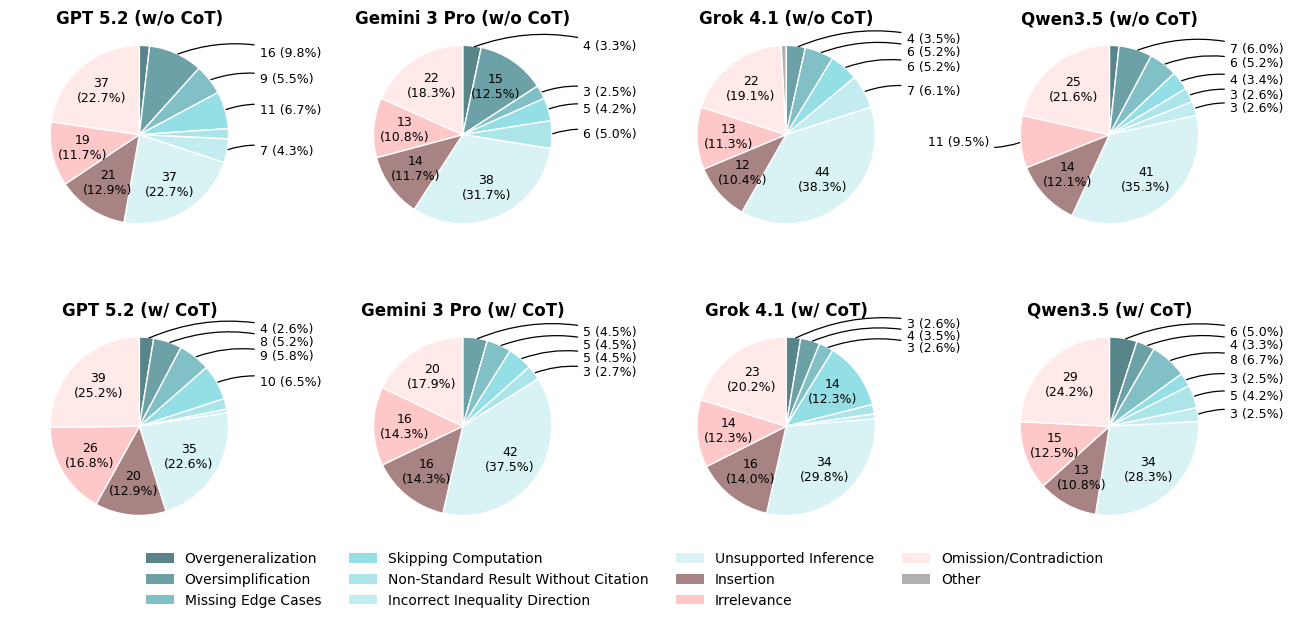}
    \caption{Distribution of the reasoning errors occurring in Setting 1.}
    \label{fig:reason-error-decomposition}
\end{figure*}

\begin{table*}[t]%
\centering%
\scriptsize%
\setlength{\tabcolsep}{3.2pt}%
\renewcommand{\arraystretch}{1.15}%
\caption{Total counts of error types made by the evaluated LLMs under each evaluation setting.}%
\label{tab:error_breakdown_settings}%
\begin{adjustbox}{max width=0.95\textwidth}%
\begin{tabular}{l*{13}{c}}%
\toprule%
\makecell{Evaluation Setting} &
\makecell{Overgene-\\ralization} &
\makecell{Oversim-\\plification} &
\makecell{Missing\\Edge Cases} &
\makecell{Skipping\\Computation} &
\makecell{Non-Standard Result\\Without Citation} &
\makecell{Incorrect Inequality\\Direction} &
\makecell{Unsupported\\Inference} &
\makecell{\textbf{Logic Error} \\\textbf{Total}$\downarrow$} &
Insertion &
Irrelevance &
\makecell{Omission/\\Contradiction} &
\makecell{\textbf{Hallucination}\\\textbf{Total}$\downarrow$} &
Other \\
\midrule%
Setting 1 (w/o CoT) & 9  & 42 & 24 & 26 & 12 & 17 & 160 & 272 & 61 & 56 & 106 & 216 & 1 \\
Setting 1 (w/ CoT)  & 13 & 21 & 25 & 32 & 13 & 5  & 145 & 239 & 65 & 71 & 111 & 244 & 1 \\
Setting 2 (w/o CoT) & 14 & 31 & 19 & 25 & 8  & 7  & 136 & 230 & 45 & 31 & 82  & 154 & 0 \\
Setting 2 (w/ CoT)  & 6  & 43 & 27 & 26 & 7  & 14 & 142 & 255 & 44 & 36 & 84  & 162 & 1 \\
Setting 3 (w/o CoT) & 12 & 22 & 32 & 43 & 22 & 7  & 166 & 285 & 57 & 19 & 69  & 137 & 4 \\
Setting 3 (w/ CoT)  & 16 & 32 & 37 & 31 & 21 & 15 & 172 & 293 & 53 & 13 & 71  & 129 & 1 \\
Setting 4 (w/o CoT) & 11 & 24 & 29 & 28 & 8  & 9  & 129 & 222 & 45 & 9  & 57  & 103 & 2 \\
Setting 4 (w/ CoT)  & 15 & 16 & 17 & 22 & 7  & 10 & 123 & 198 & 33 & 5  & 46  & 81  & 1 \\
\bottomrule%
\end{tabular}%
\end{adjustbox}%
\end{table*}%

\begin{table*}[ht]%
\centering%
\caption{Average First Error Place (\%): the average position at which the first error appears in the generated proof. Values are reported as percentages of the proof length (higher means the first error occurs later).}%
\label{tab:err_place}%
\resizebox{0.95\textwidth}{!}{%
\begin{tabular}{lcccccccc}
\toprule
\makecell{\textbf{Evaluation Setting}}
& \makecell{\textbf{GPT 5.2}\\\textbf{(w/o CoT)}}%
& \makecell{\textbf{GPT 5.2}\\\textbf{(w/ CoT)}}%
& \makecell{\textbf{Gemini 3 Pro}\\\textbf{(w/o CoT)}}%
& \makecell{\textbf{Gemini 3 Pro}\\\textbf{(w/ CoT)}}%
& \makecell{\textbf{Grok 4.1}\\\textbf{(w/o CoT)}}%
& \makecell{\textbf{Grok 4.1}\\\textbf{(w/ CoT)}}%
& \makecell{\textbf{Qwen3.5}\\\textbf{(w/o CoT)}}%
& \makecell{\textbf{Qwen3.5}\\\textbf{(w/ CoT)}}\\%
\midrule%
Setting 1 (no-context) & 17.93 & 22.73 & 25.07 & 25.35 & 24.41 & 24.77 & 19.53 & \textbf{29.66} \\%
Setting 2 (with context) & 24.55 & 21.98 & 22.50 & \textbf{36.64} & 31.22 & 32.14 & 36.02 & 32.38 \\%
Setting 3 (with oracle ratio) & 20.98 & 15.33 & 33.17 & \textbf{35.97} & 27.76 & 25.05 & 34.87 & 31.47 \\%
Setting 4 (context + oracle ratio) & 27.53 & 18.83 & 35.89 & 30.88 & 35.89 & 26.32 & \textbf{43.34} & 32.25 \\%
\bottomrule%
\end{tabular}%
}%
\end{table*}%

\subsection{Error Analysis.} \label{subsec:error-analysis}
To understand where reasoning breaks down, we developed an error analysis pipeline to attribute the reasoning errors in an unsuccessful  proof. We define two classes of errors in the model's reasoning process: 1) \textbf{\textit{logical errors}}, which reflect breakdowns within the reasoning chain itself,  including \textit{a)} {overgeneralization},  \textit{b)} {oversimplification},  \textit{c)} {missing edge cases},  \textit{d)} {skipped computations},  \textit{e)} {uncited non-standard results},  \textit{f)}{ incorrect inequality directions}, and  \textit{g)} {unsupported inferences}, 
and 2) \textbf{\textit{hallucination errors}}, which may be manifested as  \textit{a)} {unsupported inserted assumptions},  \textit{b)} reasoning about an {irrelevant or narrower problem}, and  \textit{c)} {omission or contradiction of given constraints}. Note that \textit{multiple errors may exist in one LLM-generated proof}.

\noindent \textbf{\textit{What types of mistakes do LLMs typically make?}} Without external knowledge (Setting-1), logical errors are the dominant failure mode across all models, with \textit{unsupported inference} being the most frequent error type (\eg\ an inference error of the form ``since A, therefore B'', where A is not sufficient to establish B). Figure \ref{fig:reason-error-decomposition} overviews the distribution of proof errors across LLM models \ZB{for Setting-1}.

\noindent \textbf{\textit{Can these mistakes be mitigated, and how?}} 
As shown in Table \ref{tab:error_breakdown_settings}, Setting-1, introducing generic CoT instructions partially reduces logical errors, yet simultaneously increases hallucination rates. This indicates that a domain-agnostic reasoning template is insufficient to anchor proofs without in-domain knowledge. In contrast, task-specific in-context lemmas curated by human researchers (Setting-2) yield dramatic improvements, which reduced both logical errors and hallucinations (Table \ref{tab:scores}). Furthermore, posterior information alone without in-context lemma (Setting-3) provides additional but secondary benefit. Taken together, these results suggest that solving rigorous robotic planning algorithmic proof tasks fundamentally requires in-domain expertise, where the relevance and quality of in-context knowledge matters far more than prompting strategy alone. \ZB{Consistent with our evaluation section, combining posterior information with supporting lemmas (Setting-4) further improves proof quality over Setting-3, suggesting that the ground-truth ratio is most useful when it is grounded by relevant in-domain lemmas rather than presented in isolation.}

\noindent \textbf{\textit{Does proof quality correlate with where errors first appear?}} Yes. For proofs of overall lower \ZB{reasoning} quality, errors tend to emerge earlier in the reasoning chain. As shown in Table~\ref{tab:err_place}, we tracked the relative position of the first error in each LLM-generated proof on a normalized scale from 0\% to 100\%. \TBD{We reported the average reasoning score} and the first error position of each setting across models and discovered that the overall score of LLM proof is positively correlated with the first-error position (Table \ref{tab:quality_position-avg}). This suggests that early errors tend to propagate downstream and compound into incorrect deductions. Notably, providing in-context lemmas not only reduces the frequency of logical errors, but also pushes the point of first error later into the proof. This implies that  domain-relevant knowledge helps LLMs sustain coherent reasoning for longer before going astray.

\begin{table}[t]
\vspace{0.1in}
\centering 
\caption{Average Reasoning score vs. First Error Position.}\vspace{-0.1in}
\label{tab:quality_position-avg}
\resizebox{\linewidth}{!}{%
\begin{tabular}{lcc}
\toprule
\textbf{Evaluation Setting} &
\textbf{Average} &
\textbf{Average} \\
&
\textbf{Reasoning Score $\uparrow$} &
\textbf{First Error Position (\%)} \\
\midrule
Setting 1 (no-context)      & 3.65 & 23.68 \\
Setting 2 (with context)    & 4.86 & 29.68 \\
Setting 3 (with oracle ratio)    & 4.30 & 28.08 \\
Setting 4 (context + oracle ratio) & 5.29 & 31.37 \\
\bottomrule
\end{tabular}%
}\vspace{-0.2in}
\end{table}

\subsection{Structural Failure and Success Patterns in LLM Generated Proofs} Beyond the quantitative counts in Table~II, qualitative inspection reveals recurring structural reasoning patterns that explain both failure and success:

Failures often stem from \textit{misinterpreting the problem definition} (\Judy{\ie, Oversimplification in Table \ref{tab:error_breakdown_settings}}), including conflating related routing variants (\eg, implicitly reducing a constrained formulation to classical TSP), misusing notation, or assuming structural properties such as completeness or metric conditions that are not guaranteed. 
Such misunderstandings typically occur early and propagate, leading to invalid bounds. A second failure mode concerns improper application of provided lemmas (\ie, unsupported inference or overgeneralization in Table \ref{tab:error_breakdown_settings}): models frequently invoke relevant propositions without verifying their preconditions. For instance, a proposition~\cite{wei2018coverage} may be treated as implying that each consecutive segment contributes at least $B/2$ new distinct cells, leading to an unjustified aggregation such as $N=\sum_j C_j$. This reflects surface-level lemma matching without condition-aware validation.

In contrast, LLMs usually succeed when the proof admits a clear modular decomposition aligned with the algorithmic structure. For example, in contour-connectivity–based coverage planning, \Judy{the correct reasoning} decomposes the environment into contour-connected components and then applies established properties to each component. When intermediate guarantees are explicitly stated and assumption-consistent, LLMs can correctly compose them into a valid global approximation bound. These results suggest that performance depends less on algebraic manipulation and more on the clarity of structural abstraction and the compatibility between \Judy{(provided in-context)} lemmas.

\section{Conclusion \& Future Work} \vspace{-0.05in}
We introduced the first benchmark that evaluates the state-of-the-art LLMs' ability to produce \emph{research-level} approximation-ratio proofs for robotic path-planning algorithms. 
Across \TBD{34} proof tasks curated from peer-reviewed robotics papers, our experiments show that even state-of-the-art models struggle to generate fully valid proofs when given only problem and algorithm descriptions. In contrast, providing \emph{task-specific in-context lemmas} consistently improves both reasoning correctness and proof relevance, outperforming generic chain-of-thought prompting and posterior access to the ground-truth ratio. Moreover, our results suggest that providing an oracle posterior information is most useful \emph{after} the model has been anchored by domain lemmas, acting as a consistency target that stabilizes the proof trajectory.
%
Finally, we presented fine-grained error analysis that separates logical failures from hallucinations and quantifies where the first error occurs, offering actionable signals for diagnosing and improving algorithmic-proof generation.

There are several promising directions for future work. 
(1) \textit{\textbf{Geometric visualization with VLMs.}} Many planning proofs hinge on geometric or spatial invariants (\eg, visibility cones, tours in metric spaces, feasibility under charging/pickup constraints). Incorporating multimodal geometric visualization, such as automatically rendering key structures (graphs, tours, regions, cones) and allowing a vision-language model to verify or refine intermediate claims, may reduce early logical slips and improve the grounding of feasibility arguments. 
(2) \textit{\textbf{A reusable library of in-context knowledge.}} Our findings indicate that proof success depends heavily on the availability and quality of domain lemmas. A natural next step is to build a standardized, compositional library of \textit{contexts} (lemmas, proof schemas, and constraint-check templates) that can be retrieved and adapted across families of computational-geometry and robot-planning problems, enabling stronger generalization beyond paper-specific cues. 
(3) \textit{\textbf{Proof-aware training and evaluation.}} The error taxonomy and first-error-position metrics invite targeted interventions: training or decoding-time methods that explicitly penalize unsupported inferences, missing feasibility checks, and constraint contradictions, and that reward citing provided lemmas at the right step. On the evaluation side, a promising direction is to integrate an LLM-as-judge as an online verifier during proof generation: the model periodically externalizes key intermediate subclaims, which are then checked for validity and support. When a subclaim is flagged, the generator can revise the local step before continuing, reducing error propagation and enabling more reliable long-horizon deductive chains. This also opens the door to subclaim-level supervision signals and evaluation, aligning training objectives with the specific failure modes observed in our benchmark.



\bibliographystyle{IEEEtran} 
 \bibliography{ref}



\end{document}